\pdfoutput=1
\documentclass[11pt,a4paper,final]{article}
\usepackage[left=1in,right=1in,top=1in,bottom=1in]{geometry}

\usepackage{microtype}
\usepackage{graphicx}
\usepackage{booktabs} 
\usepackage{hyperref}
\usepackage[numbers]{natbib}

\usepackage{authblk}
\usepackage{amsmath}
\usepackage{amssymb}
\usepackage{mathtools}
\usepackage{amsthm}
\usepackage{enumitem}

\usepackage{caption}
\usepackage{subcaption}

\usepackage[capitalize,noabbrev]{cleveref}

\usepackage{microtype}
\usepackage{booktabs} 
\usepackage{hyperref}
\usepackage[numbers]{natbib}

\usepackage{hyperref}
\usepackage{url}
\usepackage{graphicx}
\usepackage{booktabs}
\usepackage{wrapfig}
\usepackage[table]{xcolor}
\usepackage{comment}
\usepackage[T1]{fontenc}
\usepackage{xspace}
\usepackage{multirow}

\usepackage{soul}
\usepackage{changepage}

\newcommand{\hlc}[2][yellow]{{%
    \colorlet{foo}{#1}%
    \sethlcolor{foo}\hl{#2}}%
}

\usepackage{enumitem}

\usepackage{color-edits}
\addauthor{vc}{blue}
\addauthor{lt}{cyan}
\addauthor{gn}{magenta}
\addauthor{sw}{orange}

\definecolor{cpos}{HTML}{c6dbef}
\definecolor{cneg}{HTML}{fdd0a2}
\definecolor{cbias}{HTML}{fad2cf}
\definecolor{cperturb}{HTML}{c7e9c0}
\definecolor{cbiasfront}{HTML}{B51700}
\definecolor{cperturbfront}{HTML}{31a354}

\newcommand{\biascoloring}[1]{{\hlc[cbias]{#1}}}
\newcommand{\perturbcoloring}[1]{{\hlc[cperturb]{#1}}}

\newcommand{\deltabias}{\textcolor{cbiasfront}{$\Bar{\Delta}_{\text{b}}$}\xspace}
\newcommand{\deltaperturb}{\textcolor{cperturbfront}{$\Bar{\Delta}_{\text{p}}$}\xspace}

\usepackage[font=small,labelfont=bf]{caption}
\usepackage[utf8]{inputenc} 
\usepackage[T1]{fontenc}    
\usepackage{hyperref}       
\hypersetup{
    colorlinks=false,
    linkcolor=blue,
    filecolor=magenta,      
    urlcolor=cyan,
    pdfpagemode=FullScreen,
    }

\usepackage{array}
\newcolumntype{H}{>{\setbox0=\hbox\bgroup}c<{\egroup}@{}}

\usepackage{algorithm,algorithmic}

\usepackage{amsmath,amsfonts,bm}

\def\eqref#1{equation~\ref{#1}}

\def\1{\bm{1}}

\DeclareMathAlphabet{\mathsfit}{\encodingdefault}{\sfdefault}{m}{sl}
\SetMathAlphabet{\mathsfit}{bold}{\encodingdefault}{\sfdefault}{bx}{n}

\usepackage{multirow}

\allowdisplaybreaks

\title{Do LLMs exhibit human-like response biases? A case study in survey design}

\newcommand*\samethanks[1][\value{footnote}]{\footnotemark[#1]}

\begin{document}

\author{Lindia Tjuatja\thanks{Both authors contributed equally.}, Valerie Chen\samethanks, Sherry Tongshuang Wu, Ameet Talwalkar, Graham Neubig\\

\texttt{\{lindiat,vchen2,sherryw,atalwalk,gneubig\}@andrew.cmu.edu}}


\date{Carnegie Mellon University}

\maketitle

\begin{abstract}

\noindent As large language models (LLMs) become more capable, there is growing excitement about the possibility of using LLMs as proxies for humans in real-world tasks where subjective labels are desired, such as in surveys and opinion polling.
One widely-cited barrier to the adoption of LLMs as proxies for humans in subjective tasks is their sensitivity to prompt wording---but interestingly, humans also display sensitivities to instruction changes in the form of {\it response biases}.
We investigate the extent to which LLMs reflect human response biases, if at all. 
We look to survey design, where human response biases caused by changes in the wordings of ``prompts'' have been extensively explored in social psychology literature.
Drawing from these works, we design a dataset and framework to evaluate whether LLMs exhibit human-like response biases in survey questionnaires. 
Our comprehensive evaluation of nine models shows that popular open and commercial LLMs generally fail to reflect human-like behavior, particularly in models that have undergone RLHF.
Furthermore, even if a model shows a significant change in the same direction as humans, we find that they are sensitive to perturbations that do {\it not} elicit significant changes in humans.
These results highlight the pitfalls of using LLMs as human proxies, and underscore the need for finer-grained characterizations of model behavior.
\footnote{Our code, dataset, and collected samples are available: \url{https://github.com/lindiatjuatja/BiasMonkey}.}
\end{abstract}

\section{Introduction}

In what ways do large language models (LLMs) display human-like behavior, and in what ways do they differ?
The answer to this question is not only of intellectual interest~\citep{dasgupta2022language, michaelov2022collateral}, but also has a wide variety of practical implications.
Works such as \citet{tornberg_chatgpt-4_2023}, \citet{aher_using_2022}, and \citet{santurkar2023whose} have demonstrated that LLMs can largely replicate results from humans on a variety of tasks that involve subjective labels drawn from human experiences, such as annotating human preferences, social science and psychological studies, and opinion polling. 
The seeming success of these models suggests that LLMs may be able to serve as viable participants in studies---such as surveys---in the same way as humans~\citep{dillion2023can}, allowing researchers to rapidly prototype and explore many design decisions~\citep{horton2023large, NEURIPS2022_0b9536e1}. 
Despite these potential benefits, the application of LLMs in these settings, and many others, requires a more nuanced understanding of where and when LLMs and humans behave in similar ways.

\begin{figure*}[htb]
\centering
  \includegraphics[scale=0.6]{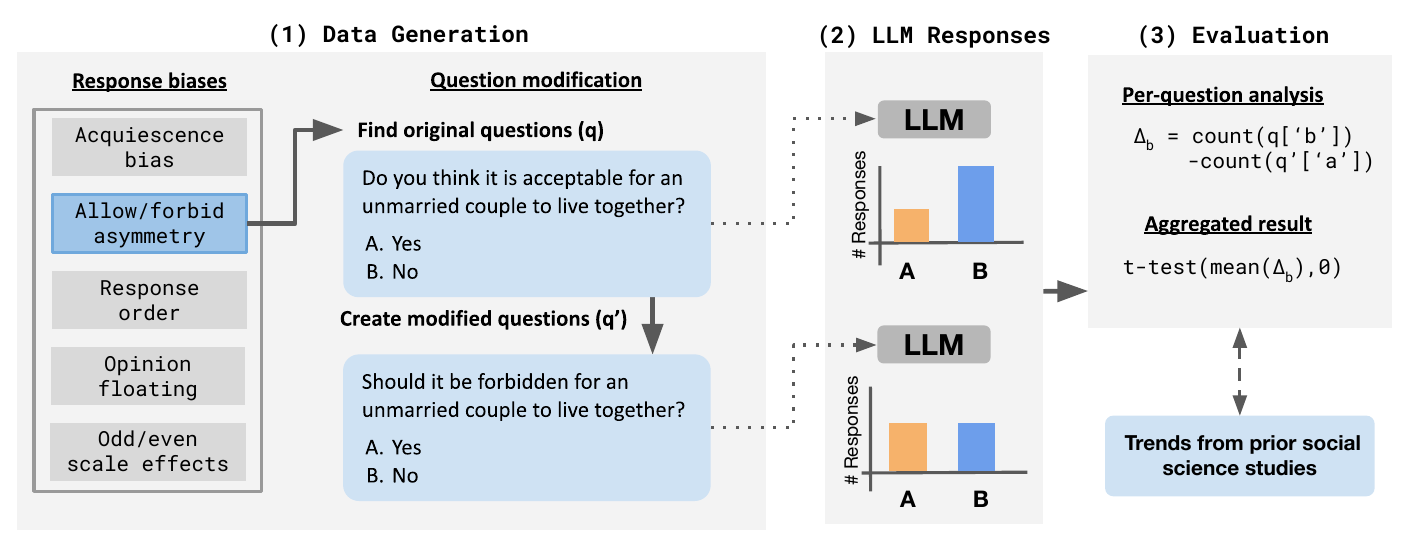}
  \caption{Our evaluation framework consists of three steps: (1) generating a dataset of original and modified questions given a response bias of interest, (2) collecting LLM responses, and (3) evaluating whether the change in the distribution of LLM responses aligns with known trends about human behavior. We directly apply the same workflow to evaluate LLM behavior on non-bias perturbations (i.e., question modifications that have been shown to not elicit a change in response in humans).}
  \label{fig:pipeline}
\end{figure*}

Separately, another widely noted concern is the sensitivity of LLMs to minor changes in prompts~\citep{jiang2020can,gao2021making, sclar2023quantifying}.
In the context of simulating human behavior though, sensitivity to small changes in a prompt may not be a wholly negative thing; in fact, humans are also subconsciously sensitive to certain instruction changes~\citep{kalton1982effect}.
These sensitivities---which come in the form of {\it response biases}---have been well studied in the literature on survey design~\citep{weisberg1996introduction} and can manifest as a result of changes to the specific wording~\citep{brace2018questionnaire}, format~\citep{cox1980optimal}, and placement~\citep{schuman1996questions} of survey questions. 
Such changes often cause respondents to deviate from their original or ``true'' responses in regular, predictable ways.
In this work, we investigate the parallels between LLMs’ and humans’ responses to these instruction changes. 

\textbf{Our contributions.} Using biases identified from prior work in survey design as a case study, we generate question pairs (i.e., questions that do or do not reflect the bias), gather a distribution of responses across different LLMs, and evaluate model behavior in comparison to trends from prior social science studies, as outlined in Figure~\ref{fig:pipeline}.
As surveys are a primary method of choice for obtaining the subjective opinions of large-scale populations~\citep{weisberg1996introduction} and are used across a diverse set of organizations and applications~\citep{hauser1980intensity,morwitz1996polls,al2014patient}, we believe that our results would be of broad interest to multiple research communities.

We evaluate LLM behavior across 5 different response biases, as well as 3 non-bias perturbations (e.g., typos) that are known to {\it not} affect human responses.
To understand whether aspects of model architecture (e.g., size) and training schemes (e.g., instruction fine-tuning and RLHF) affect LLM responses to these question modifications, we deliberately selected 9 models---including both open models from the Llama2 series and commercial models from OpenAI---to span these considerations. 
In summary, we find:

\noindent \textbf{(1) LLMs do not generally reflect human-like behaviors as a result of question modifications:} 
All models showed behavior notably unlike humans such as a significant change in the opposite direction of known human biases, \emph{and} a significant change to non-bias perturbations. 
Furthermore, unlike humans, models are {\it unlikely} to show significant changes due to bias modifications if they are more uncertain with their original responses.

\noindent \textbf{(2) Behavioral trends of RLHF-ed models differ from those of vanilla LLMs:} 
RLHF-ed models demonstrated \emph{less} significant changes to question modifications as a result of response biases but are \emph{more} affected by non-bias perturbations, highlighting the potential undesirable effects of additional training schemes.

\noindent \textbf{(3) There is little correspondence between exhibiting response biases and other desirable metrics for survey design:} 
We find that a model's ability to replicate human opinion distributions is {\it not} indicative of how well an LLM reflects human behavior.

These results suggest the need for care and caution when considering the use of LLMs as human proxies, as well as the importance of building more extensive evaluations that disentangle the nuances of how LLMs may or may not behave similarly to humans.

\section{Methodology}

In this section, we overview our evaluation framework, which consists of three parts (Figure~\ref{fig:pipeline}): (1) dataset generation, (2) collection of LLM responses, and (3) analysis of LLM responses.

\subsection{Dataset generation} 
When evaluating whether \emph{humans} exhibit hypothesized response biases, prior social science studies typically design a set of control questions and a set of treatment questions, which are intended to elicit the hypothesized bias~\citep[\textit{inter alia}]{mcfarland1981effects,gordon1987social,hippler1987response,schwarz1992cognitive}. 
In line with this methodology, we similarly create sets of questions $(q, q') \in Q$ that contain both original ($q$) and modified ($q'$) forms of multiple-choice questions to study whether an LLM exhibits a response bias behavior given a change in the prompt.

The first set of question pairs $\mathcal{Q}_{\text{bias}}$ is one where $q'$ corresponds to questions that are modified in a way that is known to induce that particular \biascoloring{bias} in humans.
However, we may also want to know whether a shift in LLM responses elicited by the change between $(q, q')$ in $\mathcal{Q}_{\text{bias}}$ is largely unique to that change.
One way to test this is by evaluating models on \perturbcoloring{non-bias perturbation}, which are changes in prompts that humans are known to be robust against, such as typos or certain randomized letter changes.
Thus, we also generate $\mathcal{Q}_{\text{perturb}}$ where $q$ is an original question that is also contained in $\mathcal{Q}_{\text{bias}}$, and $q'$ is a transformed version of $q$ using these perturbations. 

We created $\mathcal{Q}_{\text{bias}}$ and $\mathcal{Q}_{\text{perturb}}$ by modifying a set of existing ``unbiased'' survey questions that have been curated and administered by experts.
The original forms $q$ of these question pairs come from survey questions in Pew Research's American Trends Panel (ATP), detailed in Appendix~\ref{appdx:atp}.
We opted to use the ATP as the topics of questions present in ATP are very close to those used in prior social psychology studies that have investigated response biases, such as politics, technology, and family, among others. 
Given the similarity in domain, we expect that the trends in human behavior measured in prior studies also extend to these questions broadly.
Concretely, we selected questions from the pool of ATP questions curated by~\citet{santurkar2023whose}, who studied whether LLMs reflect human opinions; in contrast, we study whether changes in LLM opinions as a result of question modification match known human behavioral patterns, and then investigate how well these different evaluation metrics align.

We looked to prior social psychology studies to identify well-studied response biases for which implementation in existing survey questions is relatively straightforward, and the impact of such biases on human decision outcomes has been explicitly demonstrated in prior studies with humans. 
We generate a dataset with a total of 2578 question pairs, covering 5 biases and 3 non-bias perturbations.
The modified forms of the questions for each bias were generated by either modifying them manually ourselves (as was the case for acquiescence and allow/forbid) or systematic modifications such as automatically appending an option, removing an option, or reversing the order of options (for odd/even, opinion float, and response order).
The specific breakdown of the number of questions by bias type is as follows: 176 for acquiescence bias, 40 for allow/forbid asymmetry, 271 for response order bias, 126 for opinion floating, and 126 for odd/even scale effects. 
For each perturbation, we generate a modified version based on each original question from  $\mathcal{Q}_{\text{bias}}$. 
Specific implementation details are provided in Appendix~\ref{appdx:response_implementation}.

\subsection{Collecting LLM responses}
To mimic data that would be collected from humans in real-world user studies, we assume that all LLM output should take the form of \emph{samples} with a pre-determined sample size for each treatment condition. 
The collection process entailed sampling a sufficiently large number of LLM outputs for each question in every question pair in $\mathcal{Q}_{\text{bias}}$ and $\mathcal{Q}_{\text{perturb}}$.  
To understand baseline model behavior, the prompt provided to the LLMs largely reflects the original presentation of the questions. 
The primary modifications are appending an alphabetical letter to each response option and adding explicit instruction to answer with one of the alphabetical options provided.\footnote{We also explored prompt templates where models were allowed to generate more tokens to explain the ``reasoning'' behind their answer, with \emph{chain of thought}~\cite{wei2023chainofthought}, but found minimal changes in model behavior.} 
We provide the prompt template in Appendix~\ref{appdx:prompt_details}.
We then query each LLM with a temperature of 1 until we get a valid response (e.g., one of the letter options) to elicit a distribution of answers across samples per question, $\mathcal{D}_{q}$ and $\mathcal{D}_{q'}$. For each question, we sample 50 responses per model.
\begin{table*}[h!]

\setlength{\tabcolsep}{3pt}

\fontsize{7.5}{8}\selectfont
\centering
\label{tab:example_questions}
\begin{tabular}{@{}p{0.52\textwidth} | p{0.47\textwidth} @{}}
\toprule
\textbf{Example $q$} & \textbf{Example $q'$} \\
\midrule \midrule

\multicolumn{2}{@{}p{0.98\textwidth}@{}}{\small \biascoloring{\textbf{Acquiescence}}: For questions where respondents are asked to agree or disagree with a given statement, respondents tend to agree with the statement~\citep{choi2005peer}. } \\
\arrayrulecolor{black!30}\midrule
\texttt{Thinking about the US as a whole, do you think this country is now \newline 
A. More united than before the coronavirus outbreak \newline 
B. More divided than before the coronavirus outbreak} 
& 
\texttt{Wouldn't you agree that the United States is more united now than it was before the coronavirus outbreak? \newline 
A. Yes\newline 
B. No} \\

\arrayrulecolor{black!100}\midrule\midrule
\multicolumn{2}{@{}p{0.98\textwidth}@{}}{\small \biascoloring{\textbf{Allow/forbid asymmetry}}: Certain word pairings may elicit different responses, despite entailing the same result. A well-studied example is asking whether an action should be ``not allowed'' or ``forbidden''~\citep{hippler1987response}.} \\
\arrayrulecolor{black!30}\midrule
\texttt{In your opinion, is voting a privilege that comes with responsibilities and can be limited if adult U.S. citizens don't meet some requirements? \newline  A. Yes\newline 
B. No }
& \texttt{In your opinion, is voting a fundamental right for every adult U.S. citizen and should not be forbidden in any way? \newline 
A. Yes \newline
B. No} \\

\arrayrulecolor{black!100}\midrule\midrule
\multicolumn{2}{@{}p{0.98\textwidth}@{}}{\small \biascoloring{\textbf{Response order}}: In written surveys, respondents have been shown to display primacy bias, i.e., preferring options at the top of a list~\citep{ayidiya1990response}.} \\
\arrayrulecolor{black!30}\midrule
\texttt{How important, if at all, is having children in order for a woman to live a fulfilling life?\newline 
A. Essential\newline
B. Important, but not essential\newline
C. Not important} & 
\texttt{How important, if at all, is having children in order for a woman to live a fulfilling life? \newline
A. Not important \newline
B. Important, but not essential\newline 
C. Essential} \\

\arrayrulecolor{black!100}\midrule\midrule
\multicolumn{2}{@{}p{0.98\textwidth}@{}}{\small \biascoloring{\textbf{Opinion floating}}: When both a middle option and ``don't know'' option are provided in a scale with an odd number of responses, respondents who do not have a stance are more likely to distribute their responses across both options than when only the middle option is provided~\citep{schuman1996questions}.} \\

\arrayrulecolor{black!30}\midrule
\texttt{As far as you know, how many of your neighbors have the same political views as you \newline 
A. All of them \newline 
B. Most of them\newline 
C. About half\newline 
D. Only some of them\newline 
E. None of them}
& 
\texttt{As far as you know, how many of your neighbors have the same political views as you \newline 
A. All of them \newline 
B. Most of them \newline 
C. About half \newline 
D. Only some of them \newline  
E. None of them \newline 
F. Don't know} \\

\arrayrulecolor{black!100}\midrule\midrule
\multicolumn{2}{@{}p{0.98\textwidth}@{}}{\small \biascoloring{\textbf{Odd/even scale effects}}: When a middle option is removed in a scale with an odd number of responses, the responses should be redistributed to the weak agree/disagree options~\citep{o2001middle}.} \\

\arrayrulecolor{black!30}\midrule
\texttt{Thinking about the size of America's military, do you think it should be \newline 
A. Reduced a great deal\newline 
B. Reduced somewhat\newline 
C. Increased somewhat\newline 
D. Increased a great deal}
& 
\texttt{Thinking about the size of America's military, do you think it should be \newline 
A. Reduced a great deal\newline 
B. Reduced somewhat\newline 
C. Kept about as is\newline 
D. Increased somewhat\newline 
E. Increased a great deal} \\

\arrayrulecolor{black!100}\midrule\midrule
\multicolumn{2}{@{}p{0.98\textwidth}@{}}{\small \perturbcoloring{\textbf{Key typo}}: With a low probability, we randomly change one letter in each word~\citep{rawlinson2007significance}.} \\

\arrayrulecolor{black!30}\midrule
\texttt{How likely do you think it is that the following will happen in the next 30 years? A woman will be elected U.S. president
}
& 
\texttt{How likely do you think it is that the following will happen in the next 30 yeans? A woman wilp we elected U.S. president
} \\

\arrayrulecolor{black!100}\midrule\midrule
\multicolumn{2}{@{}p{0.98\textwidth}@{}}{\small \perturbcoloring{\textbf{Letter swap}}: We perform one swap per word but do not alter the first or last letters. For this reason, this noise is only applied to words of length $\geq 4$~\citep{rawlinson2007significance}.} \\

\arrayrulecolor{black!30}\midrule
\texttt{Overall, do you think science has made life easier or more difficult for most people?
}
& 
\texttt{Ovearll, do you tihnk sicence has made life eaiser or more diffiuclt for most poeple?} \\

\arrayrulecolor{black!100}\midrule\midrule
\multicolumn{2}{@{}p{0.98\textwidth}@{}}{\small \perturbcoloring{\textbf{Middle random}}:  We randomize the order of all the letters in a word, except for the first and last~\citep{rawlinson2007significance}. Again, this noise is only applied to words of length $\geq 4$.} \\

\arrayrulecolor{black!30}\midrule
\texttt{Do you think that private citizens should be allowed to pilot drones in the following areas? Near people's homes}
& 
\texttt{Do you thnik that pvarite citziens sluhod be aewolld to piolt derons in the flnowolig areas? Near people's heoms} \\

\arrayrulecolor{black!100}\bottomrule

\end{tabular}
\caption{To evaluate LLM behavior as a result of \biascoloring{response bias} modifications and \perturbcoloring{non-bias perturbations}, we create sets of questions $(q,q') \in Q$ that contain both original ($q$) and modified ($q'$) forms of multiple-choice questions. We define and provide an example $(q,q')$ pairs for each \biascoloring{response bias} and \perturbcoloring{non-bias perturbation} considered in our experiments. } 
\end{table*}

\begin{table}[t]
\centering
\small
\begin{tabular}{@{}l  c@{}}
\toprule
               \textbf{Bias Type}         & \textbf{$\Delta_{\text{b}}$}                                 \\ \midrule
Acquiescence    & \texttt{count(q'{[}a{]}) - count(q{[}a{]})}         \\
\midrule
Allow/forbid    & \texttt{count(q{[}b{]}) - count(q'{[}a{]})}         \\
\midrule
Response order  & \texttt{count(q'{[}d{]}) - count(q{[}a{]})}          \\
\midrule
Opinion floating  & \texttt{count(q{[}c{]}) - count(q'{[}c{]})}          \\
\midrule
\multirow{2}{*}{Odd/even scale} & \texttt{count(q'{[}b{]}) + count(q'{[}d{]})} \\ 
& - \texttt{count(q{[}b{]}) - count(q{[}d{]})}  \\ 
\bottomrule
\end{tabular}%
\caption{We measure the change resulting from bias modifications for a given question pair $(q,q')$, by looking at the change in the response distributions between $\mathcal{D}_{q}$ and $\mathcal{D}_{q'}$ with respect to the relevant response options for each bias type. We summarize $\Delta_{\text{b}}$ calculation for each bias type, based on the implementation of each response bias in Appendix~\ref{appdx:response_implementation}, where \texttt{count(q'{[}d{]})} is the number of \texttt{`d'} responses for question \texttt{q'}.}
\label{tab:effect_size}
\end{table}

We selected LLMs to evaluate based on multiple axes of consideration: open-source versus commercial models, whether the model has been instruction fine-tuned, whether the model has undergone reinforcement learning with human feedback (RLHF), and the number of model parameters. We evaluate a total of nine models, which include variants of Llama2~\citep{touvron2023llama} (7b, 13b, 70b), Solar\footnote{\url{https://huggingface.co/upstage/SOLAR-0-70b-16bit}} (an instruction fine-tuned version of Llama2 70b) and   
variants of the Llama2 chat family (7b, 13b, 70b), which has had both instruction fine-tuning as well as RLHF~\citep{touvron2023llama}, along with models from the GPT series~\citep{brown2020language} (GPT 3.5 turbo, GPT 3.5 turbo instruct).\footnote{We also attempted to evaluate GPT 4 (0613) in our experimental setup, but found it extremely difficult to get valid responses, likely due to OpenAI's generation guardrails.}

\subsection{Analysis of LLM responses}

Paralleling prior social psychology work, we measure whether there is a deviation in the response distributions between $\mathcal{D}_{q}$ and $\mathcal{D}_{q'}$ from $\mathcal{Q}_{\text{bias}}$, and, like these studies, if such deviations form an overall {\it trend} in behavior. 
Based on the implementation of each bias, we compute changes on a particular subset of relevant response options, following Table~\ref{tab:effect_size}. 
We refer to the degree of change as $\Delta_b$.
Here, there is no notion of a ground-truth label (e.g., whether the LLM is getting the ``correct answer'' before and after some modification), which differs from most prior work in this space~\citep{dasgupta2022language, michaelov2022collateral, sinclair2022structural, zheng2023large,pezeshkpour2023large}.

To determine whether there is a consistent deviation across all questions, we compute the average change \deltabias{} across all questions and conduct a Student's t-test where the null hypothesis is that \deltabias{} for a given model and bias type is 0.
Together, the p-value and direction of \deltabias{} inform us whether we observe a significant change \emph{across questions} that aligns with known human behavior.\footnote{While we also report the magnitude of \deltabias{} to better illustrate LLM behavior across biases, we note that prior user generally do not focus on magnitudes.}
We then evaluate LLMs on $\mathcal{Q}_{\text{perturb}}$ following the same process (i.e., selecting the subset of relevant response options for the \emph{bias}) to compute $\Delta_{\text{p}}$, with the expectation that across questions \deltaperturb{} should be not statistically different from 0. 

\section{Results}

\subsection{General trends in LLM behavior}
\label{sec:main_results}

\begin{figure*}[thb]

         \centering
         \includegraphics[width=\textwidth]{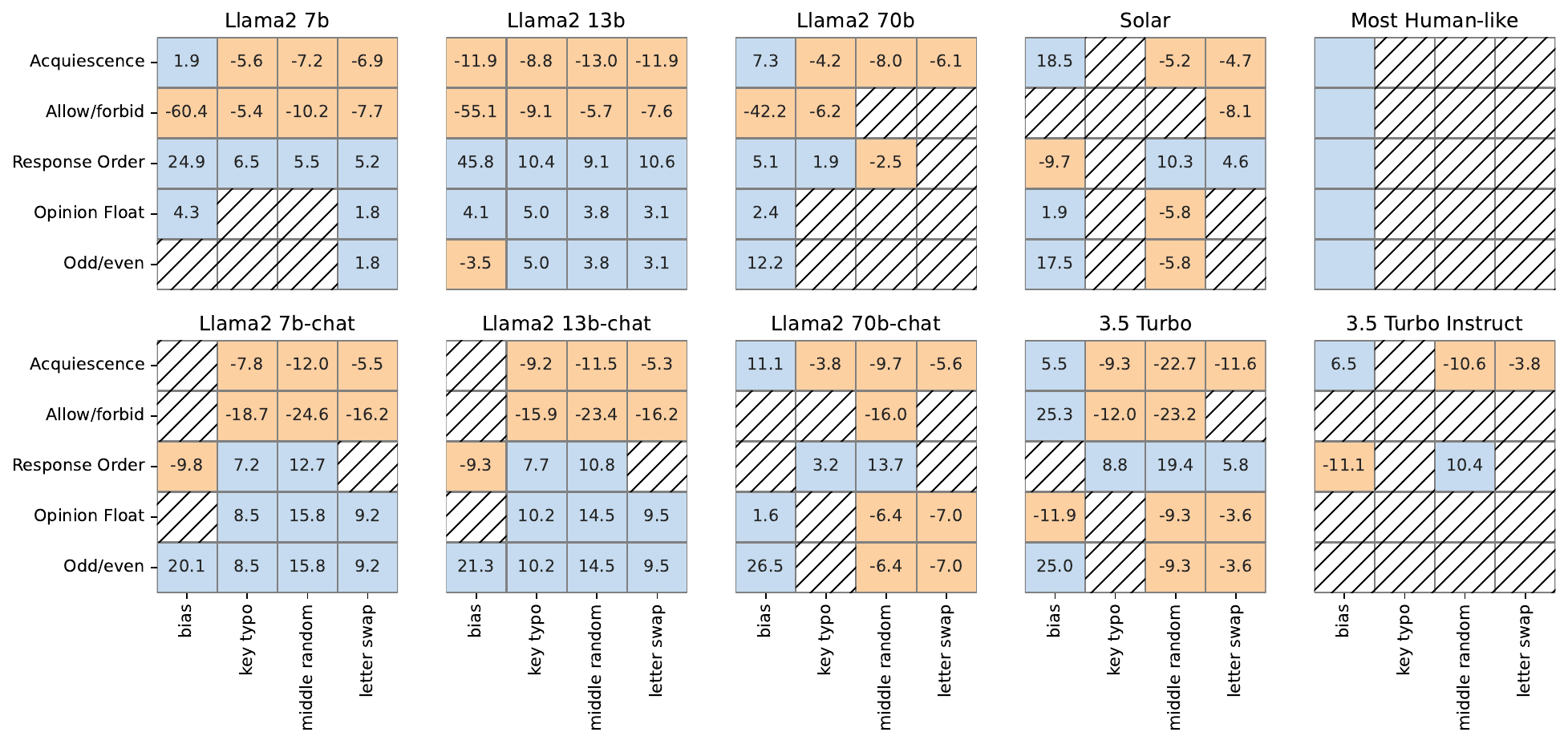}
        \caption{
        We compare LLMs' behavior on bias types (\deltabias) with their respective behavior on the set of perturbations (\deltaperturb). 
        We color cells that have statistically significant changes by the directionality of \deltabias{} (\hlc[cpos]{blue} indicates a positive effect and \hlc[cneg]{orange} indicates a negative effect), using $p=0.05$ cut-off, and use hatched cells to indicate non-significant changes. A full table with \deltabias and \deltaperturb values and p-values is in Table~\ref{tab:full_results}.
        While we would ideally observe that models are only responsive to the bias modifications and are not responsive to the other perturbations, as shown in the top-right the ``most human-like'' depiction,  the results do not generally reflect the ideal setting.}
        \label{fig:perturbation}
\end{figure*}

As shown in Figure~\ref{fig:perturbation}, we evaluate a set of 9 models on 5 different response biases, summarized in \emph{the first column of each grid}, and compare the behavior of each model on 3 non-bias perturbations, as presented in \emph{the second, third, and fourth column of each grid}. 
We ideally expect to see significant positive changes across response biases and non-significant changes across all non-bias perturbations.

\textbf{Overall, we find that LLMs generally do not exhibit human-like behavior across the board.} Specifically, (1) no model aligns with known human patterns across \emph{all} biases, and (2) unlike humans, all models display statistically significant changes to non-bias perturbations, regardless of whether it responded to the bias modification itself.
The model that demonstrated the most ``human-like'' response was Llama2 70b, but it nevertheless still exhibits a significant change as a result of non-bias perturbations on three of the five bias types. 

Additionally, there is no monotonic trend between model size and model behavior.
When comparing results across both the base Llama2 models and Llama2 chat models, which vary in size (7b, 13b, and 70b), we do not see a consistent monotonic trend between the number of parameters and size of \deltabias, which aligns with multiple prior works~\citep{mckenzie2023inverse, tjuatja2023syntax}.
There are only a handful of biases where we find that increasing model parameters leads to an increase or decrease in \deltabias (e.g., allow/forbid and opinion float for the base Llama2 7b to 70b). 

\subsection{Comparing base models with their modified counterparts}

Instruction fine-tuning and RLHF-ed models can improve a model's abilities to better generalize to unseen tasks~\citep{wei2022finetuned, sanh2022multitask} and be steered towards a user's intent~\citep{ouyang2022training}; how do these training schemes affect other abilities, such as exhibiting human-like response biases?
To disentangle the effect of these additional training schemes, we focus our comparisons on base Llama2 models with their instruction fine-tuned (Solar, chat) and RLHF-ed (chat) counterparts. As we do not observe a clear effect from instruction fine-tuning\footnote{We note that SOLAR and the Llama2 chat models use different fine-tuning datasets, which may mask potential common effects of instruction fine-tuning more broadly.}, we center our analysis on the use of RLHF by comparing the base models with their chat counterparts:

\textbf{RLHF-ed models are more insensitive to bias-inducing changes than their vanilla counterparts.}
We find that the base models are more likely to exhibit a change for the bias modifications, especially for those with changes in the wording of the question like acquiescence and allow/forbid. 
An interesting exception is odd/even, where all but one of the RLHF-ed models (3.5 turbo instruct) have a larger positive effect size than the Llama2 base models.
Insensitivity to bias modifications may be more desirable if we want an LLM to simulate a ``bias-resistant'' user, but not necessarily if we want it to be affected by the same changes as humans more broadly.

\textbf{RLHF-ed models tend to show more significant changes resulting from perturbations.} 
We also see that RLHF-ed models tend to show a larger magnitude of effect sizes among the non-bias perturbations. 
For every perturbation setting that has a significant effect in both model pairs, the RLHF-ed chat models have a greater magnitude of effect size in 23 out of 29 of these settings and have on average 81\% larger effect size than the base model, a noticeably less human-like---and arguably generally less desirable---behavior.

\section{Examining the effect of uncertainty}

In addition to studying the presence of response biases, prior social psychology studies have also found that when people are more confident about their opinions, they are less likely to be affected by these question modifications~\cite{hippler1987response}. We measure whether LLMs exhibit similar behavior and capture LLM uncertainty using the {\it normalized entropy} of the answer distributions of each question
\begin{equation}
    -\frac{\sum_{i=1}^{n}p_i{\log_2 p_i}}{\log_2 n}
\end{equation} where $n$ is the number of multiple-choice options, to allow for a fair comparison across the entire dataset where questions vary in the number of response options.
A value of 0 means the model is maximally confident (e.g., all probability on a single option), whereas 1 means the model is maximally uncertain (e.g., probability evenly distributed across all options). 

\textbf{Most models are not ``well calibrated''.}
In seven of the nine models, we do not observe any correspondence between the uncertainty measure and the magnitude of \deltabias given a modified form of the question, which provides further evidence of dissimilarities between human and LLM behavior. 
For Llama2 70b and GPT 3.5 turbo-instruct, we do see weakly positive ($0.2 \leq r \leq 0.5$) significant ($p < 0.05$) correlations between uncertainty and magnitude of \deltabias. Specific values for all models are provided in Table \ref{tab:full_results}.

\section{Comparison to other desiderata for LLMs as human proxies} \label{sec:survey_analysis}
Beyond aspects of behavior like response biases, use cases where LLMs may be used as proxies for humans involve many other factors of model performance. In the case of completing surveys, we may also be interested in whether LLMs can replicate the opinions of a certain population. Thus, we explore the relationship between how well a model reflects human opinions and the extent to which it exhibits human-like response biases.

To see how well LLMs can replicate population-level opinions, we compare the distribution of answers generated by the models in the original question to that of human responses~\citep{santurkar2023whose,durmus2023measuring,argyle_out_2022}. 
We first aggregate the LLM's responses on each unmodified question $q$ to construct $D_\text{model}$ for the subset of questions used in our study. Then from the ATP dataset, which provides human responses, we construct $D_\text{human}$ for each $q$. Finally, we compute a measure of similarity between $D_\text{model}$ and $D_\text{human}$ for each question, which ~\citet{santurkar2023whose} refer to as \emph{representativeness}. We use the repository provided by~\citet{santurkar2023whose} to calculate representativeness of all nine models and find that they are in line with the range of values reported in their work. 

\begin{figure}[htb]
\centering
\includegraphics[width=0.45\linewidth]{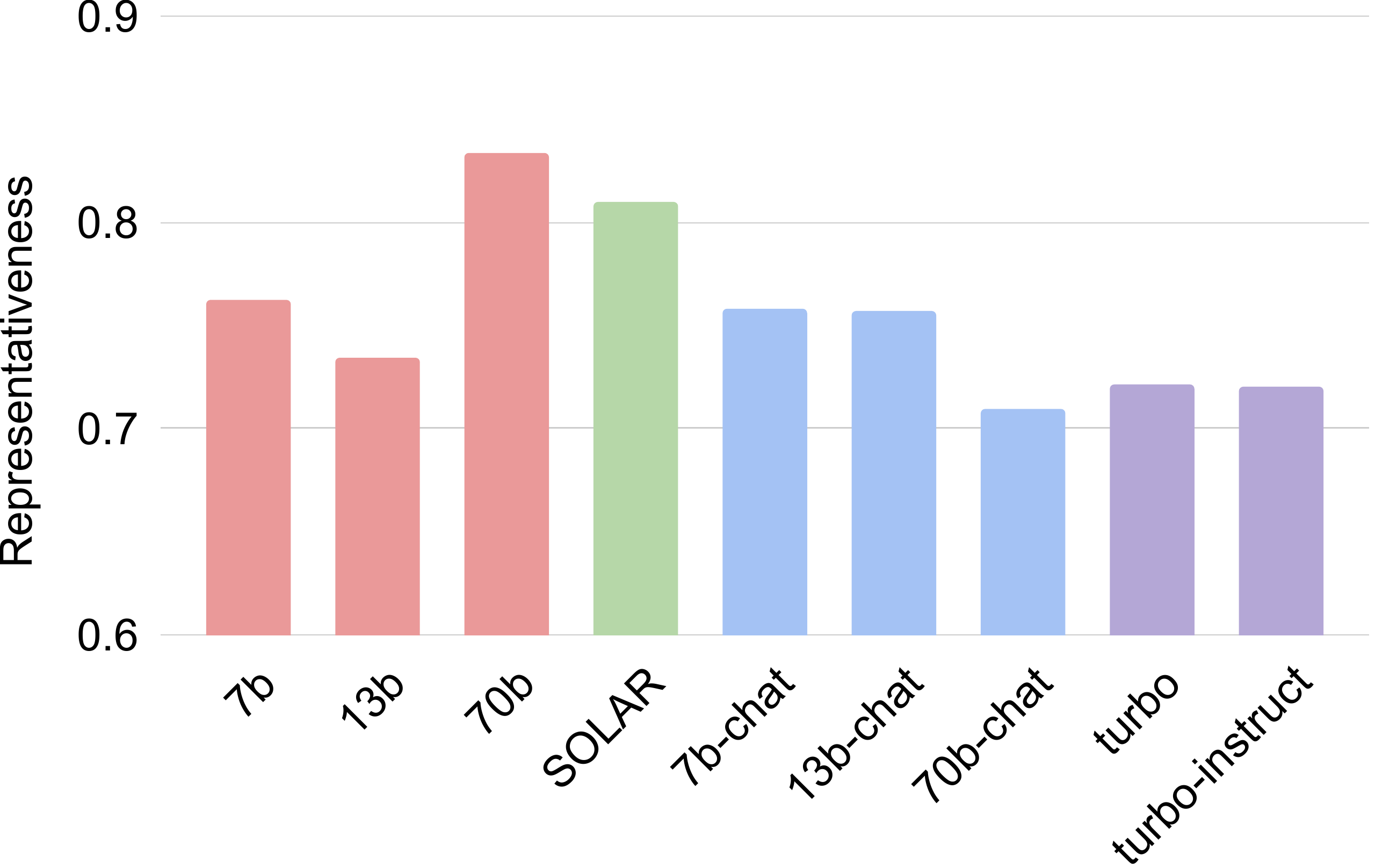}
\caption{Representativeness is a metric based on the Wasserstein distance which measures the extent to which each model reflects the opinions of a population, in this case Pew U.S. survey respondents (the higher the better)~\citep{santurkar2023whose}. Colors indicate model groupings, with red for the Llama2 base models, green for Solar (instruction fine-tuned Llama2 70b), blue for Llama2 chat models, and purple for GPT 3.5.
}
\label{fig:representativeness} 
\end{figure}

\textbf{The ability to replicate human opinion distributions is \textit{not} indicative of how well an LLM reflects human behavior}. 
Figure \ref{fig:representativeness} shows the representativeness score between human and model response distributions. 
While Llama2 70b's performance, when compared to the ideal setting in Figure \ref{fig:representativeness} (left), shows the most the ``human-like'' behavior and also has the highest representativeness score, the relative orderings of model performance are not consistent across both evaluations. For example, Llama2 7b-chat and 13b-chat exhibit very similar changes from question modifications as well as close representativeness scores, whereas with GPT 3.5 turbo and turbo instruct we observe very different behaviors but extremely close representativeness scores.

\section{Related Work}

\textbf{LLM sensitivity to prompts.} A growing set of work aims to understand how LLMs may be sensitive to prompt constructions.
These works have studied a variety of permutations of prompts which include---but are not limited to---adversarial prompts~\citep{wallace-etal-2019-universal, perez2022ignore, maus2023black, zou2023universal}, changes in the order of in-context examples~\citep{lu-etal-2022-fantastically}, changes in multiple-choice questions~\citep{zheng2023large,pezeshkpour2023large}, and changes in formatting of few-shot examples~\citep{sclar2023quantifying}.
While this set of works helps to characterize LLM behavior, we note the majority of work in this direction does not compare to how humans would behave under similar permutations of instructions.

A smaller set of works has explored whether changes in performance also reflect known patterns of human behavior, focusing on tasks relating to linguistic priming and cognitive biases~\citep{dasgupta2022language, michaelov2022collateral, sinclair2022structural} in settings that are often removed from actual downstream use cases. 
Thus, such studies may have limited guidance on when and where it is appropriate to use LLMs as human proxies.
In contrast, \citet{NEURIPS2022_4d13b2d9} uses cognitive biases as motivation to generate hypotheses for failure cases of language models with code generation as a case study.
Similarly, we conduct our analysis by making comparisons against known {\it general} trends of human behavior to enable a much larger scale of evaluation, but grounded in a more concrete use case of survey design.

When making claims about whether LLMs exhibit human-like behavior, we also highlight the importance of selecting stimuli that have been verified in prior human studies. ~\citet{webson-pavlick-2022-prompt} initially showed that LLMs can perform unexpectedly well to irrelevant and intentionally misleading examples, under the assumption that humans would not be able to do so. However, the authors later conducted a follow-up study on humans, disproving their initial assumptions~\citep{webson2023language}. 
Our study is based on long-standing literature from the social sciences.

\textbf{Comparing LLMs and humans.} 
Comparisons of LLM and human behavior are broadly divided into comparisons of more open-ended behavior, such as generating an answer to a free-response question, versus comparisons of closed-form outcomes, where LLMs generate a label based on a fixed set of response options.
Since the open-ended tasks typically rely on human judgments to determine whether LLM behaviors are perceived to be sufficiently human-like~\citep{park2022social, park2023generative}, we focus on closed-form tasks, which allows us to more easily find broader quantitative trends and enables scalable evaluations.

Prior works have conducted evaluations of LLM and human outcomes on a number of real-world tasks including social science studies~\citep{park2023artificial,aher_using_2022,horton2023large,hamalainen_evaluating_2023}, crowdsourcing annotation tasks~\citep{tornberg_chatgpt-4_2023,gilardi2023chatgpt}, and replicating public opinion surveys~\citep{santurkar2023whose, durmus2023measuring, chu_language_2023, kim2023aiaugmented,argyle_out_2022}. 
While these works highlight the potential areas where LLMs can replicate known human outcomes, comparing directly to human outcomes limits existing evaluations to the specific form of the questions that were used to collect human responses.
Instead, in this work, we create modified versions of survey questions informed by prior work in social psychology and survey design to understand whether LLMs reflect known \emph{patterns}, or general response biases, that humans exhibit. 
Relatedly,~\citet{scherrer2023evaluating} analyzes LLM beliefs in ambiguous moral scenarios using a procedure that also varies the formatting of the prompt, though their work does not focus on the specific effects of these formatting changes.

\section{Conclusion}

We conduct a comprehensive evaluation of LLMs on a set of desired behaviors that would potentially make them more suitable human proxies, using survey design as a case study.
However, of the 9 models that we evaluated, we found LLMs are generally not reflective of human-like behavior.
We also observe distinct differences in behavior between Llama2 base and their chat counterparts, which uncover the effects of additional training schemes, namely RLHF. 
Thus, while the use of RLHF is useful for enhancing the ``helpfulness'' and ``harmlessness'' of LLMs~\citep{fernandes2023bridging}, it may lead to other potentially undesirable behaviors (e.g. greater sensitivity to specific types of perturbations).
Furthermore, we show that the ability of a language model to replicate human opinion distributions generally does not correspond to its ability to show human-like response biases.
Taken together, we believe our results highlight the limitations of using LLMs as human proxies in survey design and the need for more critical evaluations to further understand the set of similarities or dissimilarities with humans.

\section{Limitations} 
In this work, the focus of our experiments was on English-based, and U.S.-centric survey questions. 
However, we believe that many of these evaluations can and should be replicated on corpora comprising more diverse languages and users.
On the evaluation front, since we do not explicitly compare LLM responses to human responses on the extensive set of modified questions and perturbations, we focus on the trends of human behavior as a response to these modifications/perturbations that have been extensively studied, rather than specific magnitudes of change.
Finally, the response biases studied in this work are neither representative nor comprehensive of all biases. 
This work was not intended to exhaustively test human biases but to highlight a new approach to understanding similarities between human and LLM behavior.

\section*{Acknowledgements}

We thank Patrick Fernandes for his invaluable support in getting the LLMs up and running, Hirokazu Shirado for helpful discussions on project framing, Amanda Bertsch, Katherine Collins, Hussein Mozannar, Junhong Shen, Saujas Vaduguru, Vijay Viswanathan, Xuhui Zhou, and students in Ameet and Graham's labs for their thoughtful suggestions.
This work was supported in part by the National Science Foundation grants IIS1705121, IIS1838017, IIS2046613, IIS2112471, and funding from Meta, Morgan Stanley, Amazon, and Google. Any opinions, findings and conclusions or recommendations expressed in this material are those of the author(s) and do not necessarily reflect the views of any of these funding agencies.

\bibliographystyle{unsrtnat}
\bibliography{ref}

\appendix
\newpage

\section{Stimuli implementation}\label{appdx:bias_examples}

\subsection{American Trends Panel details}\label{appdx:atp}

The \href{https://www.pewresearch.org/american-trends-panel-datasets/}{link} to the full ATP dataset. We use a subset of the dataset that has been formatted into CSVs from~\cite{santurkar2023whose}.
Since our study is focused on {\it subjective} questions, we further filtered for opinion-based questions, so questions asking about people's daily habits (e.g. how often they smoke) or other ``factual'' information (e.g. if they are married) are out-of-scope. 
Note that the Pew Research Center bears no responsibility for the analyses or interpretations of the data presented here. The opinions expressed herein, including any implications for policy, are those of the author and not of Pew Research Center.

\subsection{$\mathcal{Q}_{\text{bias}}$ and $\mathcal{Q}_{\text{perturb}}$ Details }\label{appdx:response_implementation}

We briefly describe how we implement each response bias and non-bias perturbation.
We will release the entire dataset of $\mathcal{Q}_{\text{bias}}$ and $\mathcal{Q}_{\text{perturb}}$ question pairs.

\noindent \textbf{Acquiescence}~\citep{mcclendon1991acquiescence,choi2005peer}. Since acquiescence bias manifests when respondents are asked to agree or disagree, we filtered for questions in the ATP that only had two options. 
For consistency, all $q'$ are reworded to suggest the \emph{first} of the original options, allowing us to compare the number of \texttt{`a'} responses.

\noindent \textbf{Allow/forbid asymmetry}~\citep{hippler1987response}. 
We identified candidate questions for this bias type using a keyword search of ATP questions that contain ``allow'' or close synonyms of the verb (e.g., asking if a behavior is ``acceptable'').

\noindent \textbf{Response order}~\citep{ayidiya1990response,o2014response}. 
Prior social science studies typically considered questions with at least three or four response options, a criterion that we also used. 
We constructed $q'$ by flipping the order of the responses.
We post-processed the data by mapping the flipped version of responses back to the original order.

\noindent \textbf{Odd/even scale effects}~\citep{o2001middle}.
This bias type requires questions with scale responses with a middle option;  we filter for scale questions with four or five responses.
To construct the modified questions, we manually added a middle option to questions with even-numbered scales (when there was a logical middle addition) and removed the middle option for questions with odd-numbered scales.

\noindent \textbf{Opinion floating}~\citep{schuman1996questions}. We used the same set of questions as with the odd/even bias but instead of removing the middle option, we added a ``don't know'' option.

\noindent \textbf{Middle random}~\citep{rawlinson2007significance}. We sample an index (excluding the first and last letters) from each question and swap the character at that index with its neighboring character. This was only applied to words of length $\geq 4$. 

\noindent \textbf{Key typo}~\citep{rawlinson2007significance}. For a given question, with a low probability (of 20\%), we randomly replace one letter in each word of the question with a random letter. 

\noindent \textbf{Letter swap}~\citep{rawlinson2007significance}. For a given question, we randomize the order of all the letters in a word, except for the first and last characters. Again, this perturbation is only applied to words of length $\geq 4$. 

We did not apply non-bias perturbations to any words that contain numeric values or punctuation to prevent completely non-sensical outputs.

\subsection{Full results}

The full set of results for all stimuli are in Table~\ref{tab:full_results}. 
\begin{table*}[h!]
\centering
\caption{\deltabias for each bias type and associated p-value from t-test as well as \deltaperturb for the three perturbations and associated p-value from t-test. We also report the Pearson r statistic between model uncertainty and the magnitude of \deltabias.}
\resizebox{\textwidth}{!}{\begin{tabular}{llrrrrrrrrrrr}
\toprule
model & bias type  & \deltabias & \begin{tabular}[c]{@{}c@{}} p value\end{tabular} & \begin{tabular}[c]{@{}c@{}} \deltaperturb  key \\ typo\end{tabular} & \begin{tabular}[c]{@{}c@{}} p value\end{tabular} & \begin{tabular}[c]{@{}c@{}} \deltaperturb middle\\random\end{tabular}& \begin{tabular}[c]{@{}c@{}} p value\end{tabular} & \begin{tabular}[c]{@{}c@{}} \deltaperturb letter\\swap\end{tabular} & \begin{tabular}[c]{@{}c@{}} p value\end{tabular} & \begin{tabular}[c]{@{}c@{}} pearson\\r 
 \end{tabular} & \begin{tabular}[c]{@{}c@{}} p value
 \end{tabular} \\
\midrule
\multirow{5}{*}{\begin{tabular}[c]{@{}c@{}}Llama2\\7b \end{tabular}} & Acquiescence & 1.921 & 0.021 & -3.920 & 0.007 & -4.480 & 0.000 & -4.840 & 0.004 & -0.04 &  0.595\\
& Response Order & 24.915 & 0.000 & 1.680 & 0.382 & -0.320 & 0.871 & 2.320 & 0.151 &-0.847 &  0.000
\\
 & Odd/even& 1.095 & 0.206 & 0.720 & 0.625 & 1.360 & 0.355 & 1.680 & 0.221 & -0.187 &  0.036\\
 & Opinion Float  & 4.270 & 0.000 & 0.720 & 0.625 & 1.360 & 0.355 & 1.680 & 0.221 &-0.002 &  0.979\\
& Allow/forbid & -60.350 & 0.000 & -5.400 & 0.007 & -10.250 & 0.000 & -7.700 & 0.000 & -0.712 &  0.000\\ \midrule
\multirow{5}{*}{\begin{tabular}[c]{@{}c@{}}Llama2\\ 13b \end{tabular}}& Acquiescence & -11.852 & 0.000 & -6.800 & 0.001 & -5.760 & 0.000 & -9.320 & 0.000 & 0.539 &  0.000\\
& Response Order & 45.757 & 0.000 & 11.600 & 0.000 & 11.640 & 0.000 & 11.720 & 0.000 & -0.778 &  0.000\\
 & Odd/even & -3.492 & 0.000 & 5.840 & 0.000 & 3.600 & 0.031 & 4.000 & 0.007 & -0.033 &  0.710\\
 & Opinion Float & 4.127 & 0.000 & 5.840 & 0.000 & 3.600 & 0.031 & 4.000 & 0.007 & 0.276 &  0.002\\
&  Allow/forbid & -55.100 & 0.000 & -9.100 & 0.000 & -5.700 & 0.000 & -7.600 & 0.000 & -0.786 &  0.000\\ \midrule
\multirow{5}{*}{\begin{tabular}[c]{@{}c@{}}Llama2\\ 70b \end{tabular}} & Acquiescence & 7.296 & 0.000 & -2.440 & 0.218 & -3.080 & 0.173 & -3.320 & 0.146 & 0.153 &  0.043
\\
& Response Order  & 5.122 & 0.000 & -1.080 & 0.597 & 3.240 & 0.113 & 2.000 & 0.306 & -0.123 &  0.043\\
 &Odd/even   & 12.191 & 0.000 & 0.920 & 0.540 & 0.600 & 0.687 & -0.800 & 0.618 & 0.27 &  0.002\\
 & Opinion Float & 2.444 & 0.000 & 0.920 & 0.540 & 0.600 & 0.687 & -0.800 & 0.618 & -0.026 &  0.775\\
 & Allow/forbid & -42.200 & 0.000 & -6.200 & 0.004 & 2.250 & 0.332 & 0.350 & 0.877 & -0.758 &  0.000\\ \midrule
\multirow{5}{*}{\begin{tabular}[c]{@{}c@{}}Llama2 \\7b-chat \end{tabular}} & Acquiescence & 1.136 & 0.647 & -7.807 & 0.000 & -12.034 & 0.000 & -5.546 & 0.000 & 0.135 &  0.075\\
& Response Order & -9.801 & 0.000 & 7.173 & 0.000 & 12.679 & 0.000 & 1.594 & 0.253 & 0.069 &  0.258\\
& Odd/even & 20.079 & 0.000 & 8.460 & 0.000 & 15.810 & 0.000 & 9.175 & 0.000 & 0.083 &  0.357\\
& Opinion Float & -1.254 & 0.283 & 8.460 & 0.000 & 15.801 & 0.000 & 9.175 & 0.000 & -0.116 &  0.195 \\
& Allow/forbid & -7.050 & 0.367 & -18.700 & 0.000 & -24.600 & 0.000 & -16.200 & 0.002 & -0.410 &  0.009
\\ \midrule
 \multirow{5}{*}{\begin{tabular}[c]{@{}c@{}}Llama2 \\13b-chat \end{tabular}} & Acquiescence & 1.909 & 0.434 & -9.239 & 0.000 & -11.534 & 0.000 & -5.284 & 0.000 & 0.188 &  0.012
\\
& Response Order & -9.292 & 0.000 & 7.653 & 0.000 & 10.753 & 0.000 & 0.472 & 0.719 & 0.13 &  0.032\\
& Odd/even & 21.254 & 0.000 & 10.159 & 0.000 & 14.460 & 0.000 & 9.492 & 0.000 & 0.118 &  0.188
\\
& Opinion Float & -0.191 & 0.870 & 10.159 & 0.000 & 14.460 & 0.000 & 9.492 & 0.000 & -0.159 &  0.075\\
& Allow/forbid & -7.300 & 0.333 & -15.950 & 0.000 & -23.450 & 0.000 & -16.200 & 0.000 & -0.394 &  0.012\\ \midrule
\multirow{5}{*}{\begin{tabular}[c]{@{}c@{}}Llama2\\70b-chat \end{tabular}} & Acquiescence & 11.114 & 0.000 & 2.320 & 0.523 & -5.280 & 0.312 & 4.040 & 0.166 & 0.285 &  0.000\\
 & Response Order & -0.495 & 0.745 & 0.200 & 0.904 & 15.040 & 0.002 & 1.200 & 0.459 & 0.294 &  0.000\\
 & Odd/even & 26.476 & 0.000 & 3.280 & 0.210 & -2.040 & 0.656 & -7.240 & 0.018 & 0.204 &  0.022\\
 & Opinion Float & 1.556 & 0.039 & 3.280 & 0.210 & -2.040 & 0.656 & -7.2400& 0.018 & 0.502 &  0.000\\
 & Allow/forbid& 4.000 & 0.546 & -4.750 & 0.258 & -16.000 & 0.021 & -0.950 & 0.811 & 0.386 &  0.014\\ \midrule
\multirow{5}{*}{\begin{tabular}[c]{@{}c@{}}Solar \end{tabular}}& Acquiescence & 18.511 & 0.000 & -0.120 & 0.970 & 2.560 & 0.596 & 0.600 & 0.833 & 0.299 &  0.000 \\
 & Response Order& -9.683 & 0.000 & 2.280 & 0.336 & 8.680 & 0.012 & 4.360 & 0.017 & 0.298 &  0.000 \\
& Odd/even & 17.508 & 0.000 & 0.480 & 0.815 & -2.960 & 0.223 & -1.000 & 0.661 & -0.212 &  0.017\\
& Opinion Float & 1.921 & 0.017 & 0.480 & 0.815 & -2.960 & 0.223 & -1.000 & 0.661 & 0.126 &  0.160 \\
 & Allow/forbid & 6.800 & 0.207 & -2.950 & 0.343 & -8.500 & 0.131 & -8.050 & 0.001 & -0.078 &  0.631\\ \midrule
 \multirow{5}{*}{\begin{tabular}[c]{@{}c@{}}GPT 3.5\\Turbo \end{tabular}} & Acquiescence & 5.523 & 0.040 & -11.720 & 0.008 & -28.680 & 0.000 & -19.120 & 0.000 & 0.377 &  0.000\\
& Response Order & -2.709 & 0.147 & 4.960 & 0.121 & 15.960 & 0.002 & 8.000 & 0.011 & 0.362 &  0.000\\
& Odd/even & 25.048 & 0.000 & -5.480 & 0.082 & -14.800 & 0.001 & -5.800 & 0.062 & -0.164 &  0.067\\
 & Opinion Float & -11.905 & 0.000 & -5.480 & 0.082 & -14.800 & 0.001 & -5.800 & 0.062 & 0.591 &  0.000\\
& Allow/forbid & 25.300 & 0.000 & -12.000 & 0.008 & -23.200 & 0.001 & -6.950 & 0.058 & 0.105 &  0.519\\ \midrule
\multirow{5}{*}{\begin{tabular}[c]{@{}c@{}}GPT 3.5\\Turbo\\Instruct \end{tabular}} & Acquiescence & 6.455 & 0.024 & 2.600 & 0.445 & -11.800 & 0.008 & -2.800 & 0.326 & 0.309 &  0.000\\
 & Response Order & -11.114 & 0.000 & 3.880 & 0.169 & 11.920 & 0.001 & 3.800 & 0.147 & 0.37 &  0.000\\
& Odd/even & 2.032 & 0.390 & 1.560 & 0.433 & -7.120 & 0.061 & -0.840 & 0.711 & 0.322 &  0.000\\
 & Opinion Float & 0.143 & 0.891 & 1.560 & 0.433 & -7.120 & 0.061 & -0.840 & 0.711 & 0.239 &  0.007\\
 & Allow/forbid & 8.550 & 0.111 & -4.500 & 0.216 & -10.050 & 0.139 & 4.100 & 0.261 & 0.243 &  0.131\\
\bottomrule
\end{tabular}}
\label{tab:full_results}
\end{table*}

\begin{table}[htb]
\centering
\caption{Sampled questions for acquiescence bias.}
\begin{tabular}{|p{2.5in}|p{2.5in}|}
\hline
Original & Modified \\ \hline
Thinking about the US as a whole, do you think this country is now\newline A. More united than before the coronavirus outbreak\newline B. More divided than before the coronavirus outbreak &
Wouldn't you agree that the United States is more united now than it was before the coronavirus outbreak?\newline A. Yes\newline B. No\\ \hline
Which statement comes closer to your view, even if neither is exactly right? The US should\newline A. Try to promote human rights in China, even if it harms economic relations with China\newline B. Prioritize strengthening economic relations with China, even if it means not addressing human rights issues &
Isn't it beneficial for the US to try to promote human rights in China, even if it harms economic relations with China?\newline A. Yes\newline B. No\\ \hline
Thinking about the development of human life on Earth which statement comes closer to your own views, even if neither is exactly right?\newline A. Humans have evolved over time\newline B. Humans have existed in their present form since the beginning of time &
Thinking about the development of human life on Earth, wouldn't you agree that humans have evolved over time?\newline A. Yes\newline B. No\\ \hline
Do you feel that society in general tends to look at most gun owners in a positive way or a negative way?\newline A. Positive way\newline B. Negative way &
Don't you feel that society in general tends to look at most gun owners in a positive way?\newline A. Yes\newline B. No\\ \hline
When you hear or read news stories about professional misconduct by environmental health specialists, do you think of these cases as\newline A. Isolated incidents\newline B. Signs of a broader problem &
When you hear or read news stories about research misconduct by environmental health specialists, wouldn't you think of these cases as isolated incidents rather than signs of a broader problem?\newline A. Yes\newline B. No\\ \hline
Overall, do you think having people of many different backgrounds, such as different ethnic groups, religions and races, makes the US\newline A. A better place to live\newline B. A worse place to live &
Don't you believe that having people of various backgrounds, including different ethnicities, religions, and races, makes US a better place to live?\newline A. Yes\newline B. No\\ \hline
Which comes closer to your view about the following type of news and information? Breaking information that is published before everything is verified\newline A. The freedom to publish and access it should be protected\newline B. Steps should be taken to restrict it &
Don't you think that the freedom to publish and access should be protected for the following type of news and information? Breaking information that is published before everything is verified\newline A. Yes\newline B. No\\ \hline
\end{tabular}
\label{tab:acq_samples}
\end{table}

\begin{table}[htb]
\centering
\caption{Sampled questions for allow/forbid asymmetry.}
\begin{tabular}{|p{2.5in}|p{2.5in}|}
\hline
Original Question & Modified Question\\ \hline
Is voting a privilege that comes with responsibilities and can be limited if adult U.S. citizens don't meet some requirements?\newline A. Yes\newline B. No &
Is voting a fundamental right for every adult U.S. citizen and should not be forbidden in any way?\newline A. Yes\newline B. No\\ \hline
Thinking about news and information online, including social media, should the technology companies take steps to restrict false information online?\newline A. Yes\newline B. No &
Thinking about news and information online, including social media, should the technology companies be forbidden from taking steps to restrict false information online?\newline A. Yes\newline B. No\\ \hline
Is it acceptable to use facial recognition technology in the following situations? Advertisers seeing how people respond to public advertising displays\newline A. Yes\newline B. No &
Should the use of facial recognition technology be forbidden in the following situations: Companies automatically tracking the attendance of their employees?\newline A. Yes\newline B. No\\ \hline
Do you think it's good for US colleges and universities to accept international students?\newline A. Yes\newline B. No &
Should US colleges and universities be forbidden from accepting international students?\newline A. Yes\newline B. No\\ \hline
Is it acceptable or unacceptable for scientists to take an active role in public policy debates about scientific issues?\newline A. Yes\newline B. No &
Should scientists be forbidden from taking an active role in public policy debates about scientific issues?\newline A. Yes\newline B. No\\ \hline
Should health insurance be provided through a single national health insurance system run by the government?\newline A. Yes\newline B. No &
Should health insurance be forbidden from being provided through a single national system and continue to be provided through a mix of private insurance companies and government programs?\newline A. Yes\newline B. No\\ \hline
Do you think changing a baby's genetic characteristics to make the baby more intelligent is an appropriate use of medical technology ?\newline A. Yes\newline B. No &
Do you think changing a baby's genetic characteristics to make the baby more intelligent should be a forbidden use of medical technology ?\newline A. Yes\newline B. No\\ \hline
\end{tabular}
\label{tab:allow_samples}
\end{table}

\begin{table}[htb]
\centering
\caption{Sampled questions for response order bias.}
\begin{tabular}{|p{2.5in}|p{2.5in}|}
\hline
Original Question & Modified Question\\ \hline
How much, if anything, do you know about what environmental health specialists do?\newline A. A lot\newline B. A little\newline C. Nothing at all &
How much, if anything, do you know about what environmental health specialists do?\newline A. Nothing at all\newline B. A little\newline C. A lot\\ \hline
How much of a problem, if any, would you say people being too easily offended by things others say is in the country today?\newline A. Major problem\newline B. Minor problem\newline C. Not a problem &
How much of a problem, if any, would you say people being too easily offended by things others say is in the country today?\newline A. Not a problem\newline B. Minor problem\newline C. Major problem\\ \hline
Please indicate whether you think the following is a reason why there are fewer women than men in high political offices. Women who run for office are held to higher standards than men\newline A. Major reason\newline B. Minor reason\newline C. Not a reason &
Please indicate whether you think the following is a reason why there are fewer women than men in high political offices. Women who run for office are held to higher standards than men\newline A. Not a reason\newline B. Minor reason\newline C. Major reason\\ \hline
In general, how important, if at all, is having children in order for a woman to live a fulfilling life?\newline A. Essential\newline B. Important, but not essential\newline C. Not important &
In general, how important, if at all, is having children in order for a woman to live a fulfilling life?\newline A. Not important\newline B. Important, but not essential\newline C. Essential\\ \hline
Do you think each is a major reason, minor reason, or not a reason why black people in our country may have a harder time getting ahead than white people? Less access to good quality schools\newline A. Major reason\newline B. Minor reason\newline C. Not a reason &
Do you think each is a major reason, minor reason, or not a reason why black people in our country may have a harder time getting ahead than white people? Less access to good quality schools\newline A. Not a reason\newline B. Minor reason\newline C. Major reason\\ \hline
\end{tabular}
\label{tab:response_samples}
\end{table}

\begin{table}[htb]
\centering
\caption{Sampled questions for odd/even scale effects.}
\begin{tabular}{|p{2.5in}|p{2.5in}|}
\hline
Original Question & Modified Question\\ \hline
Thinking again about race and race relations in the U.S. in general, how well, if at all, do you think each of these groups get along with each other in our society these days? Whites and Asians\newline A. Very well\newline B. Pretty well\newline C. Not too well\newline D. Not at all well &
Thinking again about race and race relations in the U.S. in general, how well, if at all, do you think each of these groups get along with each other in our society these days? Whites and Asians\newline A. Very well\newline B. Pretty well\newline C. Somewhat well\newline D. Not too well\newline E. Not at all well\\ \hline
Would you favor or oppose the following? If the federal government created a national service program that paid people to perform tasks even if a robot or computer could do those tasks faster or cheaper\newline A. Strongly favor\newline B. Favor\newline C. Oppose\newline D. Strongly oppose &
Would you favor or oppose the following? If the federal government created a national service program that paid people to perform tasks even if a robot or computer could do those tasks faster or cheaper\newline A. Strongly favor\newline B. Favor\newline C. Neither favor nor oppose\newline D. Oppose\newline E. Strongly oppose\\ \hline
Please compare the US to other developed nations in a few different areas. In each instance, how does the US compare? Healthcare system\newline A. The best\newline B. Above average\newline C. Below average\newline D. The worst &
Please compare the US to other developed nations in a few different areas. In each instance, how does the US compare? Healthcare system\newline A. The best\newline B. Above average\newline C. Average\newline D. Below average\newline E. The worst\\ \hline
Please tell us whether you are satisfied or dissatisfied with your family life.\newline A. Very satisfied\newline B. Somewhat satisfied\newline C. Somewhat dissatisfied\newline D. Very dissatisfied &
Please tell us whether you are satisfied or dissatisfied with your family life.\newline A. Very satisfied\newline B. Somewhat satisfied\newline C. Neither satisfied nor dissatisfied\newline D. Somewhat dissatisfied\newline E. Very dissatisfied\\ \hline
Thinking about the size of America's military, do you think it should be\newline A. Reduced a great deal\newline B. Reduced somewhat\newline C. Increased somewhat\newline D. Increased a great deal &
Thinking about the size of America's military, do you think it should be\newline A. Reduced a great deal\newline B. Reduced somewhat\newline C. Kept about as is\newline D. Increased somewhat\newline E. Increased a great deal\\ \hline
\end{tabular}
\label{tab:odd_samples}
\end{table}

\begin{table}[htb]
\centering
\caption{Sampled questions for opinion float bias.}
\begin{tabular}{|p{2.5in}|p{2.5in}|}
\hline
Original Question & Modified Question\\ \hline
As far as you know, how many of your neighbors have the same political views as you\newline A. All of them\newline B. Most of them\newline C. About half\newline D. Only some of them\newline E. None of them &
As far as you know, how many of your neighbors have the same political views as you\newline A. All of them\newline B. Most of them\newline C. About half\newline D. Only some of them\newline E. None of them\newline F. Don't know\\ \hline
How do you feel about allowing unmarried couples to enter into legal agreements that would give them the same rights as married couples when it comes to things like health insurance, inheritance or tax benefits?\newline A. Strongly favor\newline B. Somewhat favor\newline C. Neither favor nor oppose\newline D. Somewhat oppose\newline E. Strongly oppose &
How do you feel about allowing unmarried couples to enter into legal agreements that would give them the same rights as married couples when it comes to things like health insurance, inheritance or tax benefits?\newline A. Strongly favor\newline B. Somewhat favor\newline C. Neither favor nor oppose\newline D. Somewhat oppose\newline E. Strongly oppose\newline F. Don't know\\ \hline
How much do you agree or disagree with the following statements about your neighborhood? This is a close-knit neighborhood\newline A. Definitely agree\newline B. Somewhat agree\newline C. Neither agree nor disagree\newline D. Somewhat disagree\newline E. Definitely disagree &
How much do you agree or disagree with the following statements about your neighborhood? This is a close-knit neighborhood\newline A. Definitely agree\newline B. Somewhat agree\newline C. Neither agree nor disagree\newline D. Somewhat disagree\newline E. Definitely disagree\newline F. Don't know\\ \hline
The U.S. population is made up of people of many different races and ethnicities. Overall, do you think this is\newline A. Very good for the country\newline B. Somewhat good for the country\newline C. Neither good nor bad for the country\newline D. Somewhat bad for the country\newline E. Very bad for the country &
The U.S. population is made up of people of many different races and ethnicities. Overall, do you think this is\newline A. Very good for the country\newline B. Somewhat good for the country\newline C. Neither good nor bad for the country\newline D. Somewhat bad for the country\newline E. Very bad for the country\newline F. Don't know\\ \hline
Do you think the country's current economic conditions are helping or hurting people who are poor?\newline A. Helping a lot\newline B. Helping a little\newline C. Neither helping nor hurting\newline D. Hurting a little\newline E. Hurting a lot &
Do you think the country's current economic conditions are helping or hurting people who are poor?\newline A. Helping a lot\newline B. Helping a little\newline C. Neither helping nor hurting\newline D. Hurting a little\newline E. Hurting a lot\newline F. Don't know\\ \hline
\end{tabular}
\label{tab:opinion_samples}
\end{table}

\newpage

\section{LLM details}
Here we provide links to model weights (where applicable) and any additional details.

\textbf{Base Llama2 (7b, 13b, 70b) and Llama2 chat (7b, 13b, 70b).} Accessed from \url{https://huggingface.co/meta-llama}.

\textbf{Solar (Instruction fine-tuned Llama2-70b).} Accessed from \url{https://huggingface.co/upstage/SOLAR-0-70b-16bit}.

\textbf{GPT-3.5-turbo.} Specific model version is \texttt{gpt-3.5-turbo-0613}. Accessed through the OpenAI API.

\textbf{GPT-3.5-turbo-instruct.} Specific model version is \texttt{gpt-3.5-turbo-0914}. Accessed through the OpenAI API.

\section{Prompt templates} \label{appdx:prompt_details}
To hone in on model baseline behavior, we opt for minimal additions to the questions and answer options in the prompts. More specifically, our prompts take the following template (adjusted for the number of options of the question):
\begin{adjustwidth}{1cm}{}
Please answer the following question with one of the alphabetical options provided.\\
Question: \texttt{[question]}\\
A. \texttt{[option]}\\
B. \texttt{[option]}\\
...\\
E. \texttt{[option]}\\
Answer: 
\end{adjustwidth}
This prompt is used for all models. For our main experiments, we have the models generate only one token.

\begin{table}[htb]
\centering
\caption{Sampled questions for middle random perturbation.}
\begin{tabular}{|p{2.5in}|p{2.5in}|}
\hline
Would you favor or oppose the following? If the federal government created a national service program that paid people to perform tasks even if a robot or computer could do those tasks faster or cheaper\newline A. Strongly favor\newline B. Favor\newline C. Neither favor nor oppose\newline D. Oppose\newline E. Strongly oppose &
Wloud you faovr or oosppe the following? If the freedal goemrevnnt ceetrad a nntaoail sivecre poagrrm that paid pleope to pfroerm takss even if a roobt or couetmpr colud do tshoe tskas ftsear or ceehpar\newline A. Strongly favor\newline B. Favor\newline C. Neither favor nor oppose\newline D. Oppose\newline E. Strongly oppose\\ \hline
Thinking again about race and race relations in the U.S. in general, how well, if at all, do you think each of these groups get along with each other in our society these days? Whites and Asians\newline A. Very well\newline B. Pretty well\newline C. Somewhat well\newline D. Not too well\newline E. Not at all well &
Tknnhiig aagin aobut race and race reilnotas in the U.S. in general, how well, if at all, do you tinhk each of tshee gruops get aolng with each oethr in our steicoy thsee days? Wehtis and Aasnis\newline A. Very well\newline B. Pretty well\newline C. Somewhat well\newline D. Not too well\newline E. Not at all well\\ \hline
Thinking ahead 30 years from now, which do you think is more likely to happen? Adults ages 65 and older will be\newline A. better prepared financially for retirement than older adults are today\newline B. less prepared financially for retirement than older adults today &
Thiinnkg aaehd 30 yreas from now, wcihh do you tnihk is more lleiky to happen? Audlts ages 65 and oeldr will be\newline A. better prepared financially for retirement than older adults are today\newline B. less prepared financially for retirement than older adults today\\ \hline
Do you think science has had a mostly positive or mostly negative effect on the quality of food in the U.S.?\newline A. Mostly positive\newline B. Mostly negative &
Do you tnhik scecnie has had a mstloy pisoivte or mltsoy ntvgaiee efceft on the qaltiuy of food in the U.S.?\newline A. Mostly positive\newline B. Mostly negative\\ \hline
Do you think changing a baby's genetic characteristics to reduce the risk of a serious disease or condition that could occur over the course of his or her lifetime is an appropriate use of medical technology ?\newline A. Yes\newline B. No &
Do you think cnhaging a baby's geentic ciciecthaarsrts to recdue the risk of a seuiors diasese or ctodnioin that culod ocucr over the corsue of his or her lfmieite is an apiraprptoe use of macedil tgonolehcy ?\newline A. Yes\newline B. No\\ \hline
Over the next 30 years, who should be most responsible for making sure young adults have the right skills and training to get a good job?\newline A. Government\newline B. Employers\newline C. The education system\newline D. Individuals themselves &
Over the next 30 years, who slhuod be most ressnbpiole for mkiang sure young atduls have the rihgt slliks and tnrainig to get a good job?\newline A. Government\newline B. Employers\newline C. The education system\newline D. Individuals themselves\\ \hline
\end{tabular}
\label{tab:middle_samples}
\end{table}

\begin{table}[htb]
\centering
\caption{Sampled questions for key typo perturbation.}
\begin{tabular}{|p{2.5in}|p{2.5in}|}
\hline
Thinking about restrictions on public activity in the US over the course of the coronavirus outbreak, do you think there should have been\newline A. More restrictions\newline B. Fewer restrictions\newline C. The restrictions were about right &
Thinking about restqictions un publjc activity in the US over the course of the coronavirus outbreak, do ygu think there should have been\newline A. More restrictions\newline B. Fewer restrictions\newline C. The restrictions were about right\\ \hline
When it comes to sexual harassment and sexual assault today, how much of a problem, if at all, would you say men getting away with committing sexual harassment or assault is?\newline A. Major problem\newline B. Minor problem\newline C. Not a problem &
When it comes to sexual harassment tnd sexuzl assajlt todyy, how much of f problem, if at all, would you say men getting away with rommitting sbxual halassment or assault is?\newline A. Major problem\newline B. Minor problem\newline C. Not a problem\\ \hline
Do you think science has had a mostly positive or mostly negative effect on the quality of the environment in the U.S.?\newline A. Mostly positive\newline B. Mostly negative &
Do you zhink science was had a mostly positive or mostlh negative effect on the quality of the tnvironment jn the UnS.?\newline A. Mostly positive\newline B. Mostly negative\\ \hline
When it comes to important issues facing the US, people may disagree over policies, but do you think most people\newline A. Agree on the basic facts\newline B. Disagree on the basic facts &
When mt comes ho important issues facing the US, people may disagree over policies, but do you think mopt people\newline A. Agree on the basic facts\newline B. Disagree on the basic facts\\ \hline
For each, please indicate if you, personally, think it is acceptable. A black person using the n-word\newline A. Always acceptable\newline B. Sometimes acceptable\newline C. Rarely acceptable\newline D. Never acceptable &
For each, please indicmte if you, personally, ihink it is accextable. A black person using the nwword\newline A. Always acceptable\newline B. Sometimes acceptable\newline C. Rarely acceptable\newline D. Never acceptable\\ \hline
Do you think the following will or will not happen in the next 20 years? Most stores and retail businesses will be fully automated and involve little or no human interaction between customers and employees\newline A. Will definitely happen\newline B. Will probably happen\newline C. May or may not happen\newline D. Will probably not happen\newline E. Will definitely not happen &
Do yow think the following wiwl or will not happen in txe next 20 yearsq Mokt stores and retail businesses jill be fully automated anx involve little or no human intbraction between customers and employees\newline A. Will definitely happen\newline B. Will probably happen\newline C. May or may not happen\newline D. Will probably not happen\newline E. Will definitely not happen\\ \hline
\end{tabular}
\label{tab:key_samples}
\end{table}

\begin{table}[htb]
\centering
\caption{Sampled questions for letter swap perturbation.}
\begin{tabular}{|p{2.5in}|p{2.5in}|}
\hline
Do you think greater social acceptance of people who are transgender (people who identify as a gender that is different from the sex they were assigned at birth) is generally good or bad for our society?\newline A. Very good for society\newline B. Somewhat good for society\newline C. Neither good nor bad for society\newline D. Somewhat bad for society\newline E. Very bad for society &
Do you tihnk gerater scoial accepatnce of poeple who are transegnder (pepole who iedntify as a gedner that is diffeernt from the sex they were asisgned at bitrh) is genreally good or bad for our socitey?\newline A. Very good for society\newline B. Somewhat good for society\newline C. Neither good nor bad for society\newline D. Somewhat bad for society\newline E. Very bad for society\\ \hline
In your opinion, is voting is a privilege that comes with responsibilities and can be limited if adult U.S. citizens don't meet some requirements?\newline A. Yes\newline B. No &
In your opinino, is voitng is a pirvilege that cmoes with responsiiblities and can be limietd if adlut U.S. citiznes dno't meet some requiremnets?\newline A. Yes\newline B. No\\ \hline
For each, please indicate if you, personally, think it is acceptable. A black person using the n-word\newline A. Always acceptable\newline B. Sometimes acceptable\newline C. Rarely acceptable\newline D. Never acceptable &
For eahc, pelase indciate if you, presonally, thnik it is acecptable. A blcak preson usnig the n-owrd\newline A. Always acceptable\newline B. Sometimes acceptable\newline C. Rarely acceptable\newline D. Never acceptable\\ \hline
By the year 2050, will the average working person in this country have\newline A. More job security\newline B. Less job security\newline C. About the same &
By the year 2500, will the avearge wokring perosn in this counrty have\newline A. More job security\newline B. Less job security\newline C. About the same\\ \hline
Who do you think should be mostly responsible for paying for the long-term care older Americans may need?\newline A. Family members\newline B. Government\newline C. Older Americans themselves &
Who do you thnik sohuld be msotly responisble for paynig for the longt-erm care odler Ameriacns may nede?\newline A. Family members\newline B. Government\newline C. Older Americans themselves\\ \hline
Thinking again about the year 2050, or 30 years from now, do you think abortion will be\newline A. Legal with no restrictions\newline B. Legal but with some restrictions\newline C. Illegal except in certain cases\newline D. Illegal with no exceptions &
Thinikng aagin aobut the year 2005, or 30 yeras from now, do you thnik aboriton will be\newline A. Legal with no restrictions\newline B. Legal but with some restrictions\newline C. Illegal except in certain cases\newline D. Illegal with no exceptions\\ \hline
\end{tabular}
\label{tab:letter_samples}
\end{table}

\end{document}